\documentclass[letterpaper, 10 pt, conference]{ieeeconf}  

\IEEEoverridecommandlockouts                              
\usepackage{graphicx}

\usepackage[caption=false]{subfig}
\usepackage{xcolor}
\usepackage{array}
\usepackage{float}
\usepackage{booktabs}
\usepackage{adjustbox}
\usepackage[noadjust]{cite}
\usepackage{multirow}
\usepackage{todonotes}
\usepackage{marginnote}
\usepackage{amsmath,amssymb,amsfonts}

\usepackage[a-1b]{pdfx}

\title{Learning-based Estimation of Forward Kinematics for an \\ Orthotic Parallel Robotic Mechanism}
\author{Jingzong Zhou,$^{1,\dagger}$ Yuhan Zhu,$^{1,\dagger}$ Xiaobin Zhang,$^{1,\dagger}$ Sunil Agrawal,$^{2}$ and Konstantinos Karydis$^{1}$ %
\thanks{$^{\dagger}$ These authors contributed equally. 
$^{1}$~Dept. of Electrical and Computer Engineering, University of California, Riverside, 900 University Ave, Riverside, CA 92521, USA. Email:{\tt\footnotesize\{jzhou227, yzhu275, xzhan548, karydis\}@ucr.edu}. 
$^{2}$~Dept. of Mechanical Engineering, Columbia University, 500 W. 120th Street \#510, New York, NY 10027, USA. Email:{\tt\footnotesize sa3077@columbia.edu}.
}
}

\begin{document}
\maketitle

\begin{abstract}
This paper introduces a 3D parallel robot with three identical five-degree-of-freedom chains connected to a circular brace end-effector, aimed to serve as an assistive device for patients with cervical spondylosis. The inverse kinematics of the system is solved analytically, whereas learning-based methods are deployed to solve the forward kinematics. The methods considered herein include a Koopman operator-based approach as well as a neural network-based approach. The task is to predict the position and orientation of end-effector trajectories. The dataset used to train these methods is based on the analytical solutions derived via inverse kinematics. The methods are tested both in simulation and via physical hardware experiments with the developed robot. Results validate the suitability of deploying learning-based methods for studying parallel mechanism forward kinematics that are generally hard to resolve analytically.
\end{abstract}

\IEEEpeerreviewmaketitle

\section{Introduction} \label{Introduction}
Cervical spondylosis, a common source of neck pain caused by degenerative changes that start in the intervertebral discs, has been significantly affecting people over 30 years of age~\cite{binder2007cervical}. 
A study in testing the efficacy of intermittent cervical traction in patients with chronic neck pain has shown that $26\%$ to $71\%$ of the adult population experience an episode of neck pain in their lifetime~\cite{borman2008efficacy}. 
Although neck pain has been reported to lead to a lesser degree of disability compared to back pain, the quality of life of patients is still severely affected by chronic symptoms~\cite{popescu2020neck}. 
To mitigate the long-term impact, effective management is thus crucial. 
Typical medical management includes pharmacological and rehabilitation components, with physical therapy playing a pivotal role for the  latter~\cite{mazanec2007medical}. 
A randomized clinical study that involved 60 patients with neck pain lasting for at least six months 
revealed that patients receiving physical therapy demonstrated greater improvement in pain relief and functional outcomes than those solely with drug treatment~\cite{aslan2012effects}. 
Controlled clinical trials have substantiated that extension traction as part of cervical spine rehabilitation can improve patients' quality of life~\cite{oakley2021restoring}. 
To enhance accessibility and convenience, an assistive wearable neck brace presents a promising alternative for therapy. 
Such a device could offer patients consistent treatment and minimize the need to visit the hospital frequently.


Traction exercise neck braces, developed as a treatment for cervical spondylosis, aim to adjust the cervical vertebra structure to alleviate related neck pain~\cite{xiao2021effect}. 
Because of spatial restrictions in the head-neck region, parallel robots are particularly well-suited to deliver appropriate motion support while employing a compact structure. 
A notable example is a three-chain parallel robotic neck brace whereby each chain contains three joints (two revolute [R] and one spherical [S]) arranged in series (RRS)~\cite{zhang2019using}. 
This design provides an assist-as-needed moment around the head-neck area, ensuring both safety and efficacy in a head movement training task~\cite{zhang2019using}. 
Similarly, a study of a three degree-of-freedom (DoF) neck brace used a three-chain-RPUR parallel mechanism,\footnote{~Recall that P stands for a prismatic joint and U for a universal one.} which was developed to relieve pain in the cervical spine by delivering an upward traction force~\cite{kulkarni2024design}. 

However, operating a parallel robot can pose great challenges because of the complexity of its forward kinematics~\cite{pandilov2014comparison} and the often-times complete absence of closed-form analytical solutions. 
The intricate geometry and highly nonlinear dynamics of such robots result in non-unique solutions in most cases~\cite{prado2021artificial}. 
Although numerical solvers are commonly employed, their computational inefficiency may limit more widespread applicability, especially when in need to perform online computation on edge hardware~\cite{kang2012learning}. 
Further, the solution of forward kinematics is often attained indirectly by solving inverse kinematics first and then inverting them locally, as demonstrated in studies with a two degree-of-freedom spatial parallel mechanism~\cite{zhou2024design}. 
Motivated by these challenges, our work aims to employ data-driven methods to solve the forward kinematics based on solutions obtained from inverse kinematics. 


One data-efficient yet powerful learning-based family of methods for robotic systems is Koopman operator theory, which has been gaining attention owing to its capacity to enable real-time learning and support data-driven control~\cite{shi2024koopman}. 
The Koopman operator is not nearly as data-intensive as most other machine learning techniques and takes much less time to train, making it a suitable candidate method for hardware with lower computation capabilities and requiring a higher update frequency. 
Koopman operator theory, along with methods to estimate the operator from data (such as the Extended Dynamic Mode Decomposition with control (EDMDc)~\cite{korda2018linear}), has been applied in the modeling and control of various robotic systems~\cite{shi2023koopman}. 
Notable examples include the modeling and control of a tail-actuated robotic fish~\cite{mamakoukas2021derivative}, trajectory control for micro-aerial vehicles~\cite{shi2020data}, dynamics estimation for spherical robots~\cite{abraham2017model}, and model extraction for soft systems~\cite{bruder2019nonlinear}. 
Koopman operator theory has also been applied to the trajectory tracking control of a three-soft-fingered mechanism for zero-shot grasping~\cite{shi2022online}. 

Deep neural networks provide a promising alternative to capture the non-linear relationship between the inputs (in the joint space) and the outputs (in the cartesian space) in forward kinematics~\cite{prado2021artificial,kumar2013forward,geng1991neural}. 
The depth of deep neural networks can be flexibly adjusted, allowing different layers to capture varying levels of complexity in forward kinematics~\cite{janik2020complexity}. 
Some works have focused on deploying neural network-based methods for estimating the forward kinematics of various parallel mechanisms. 
For instance, Dehghani et al. proposed to solve the forward kinematics of parallel robot HEXA with a multilayer perceptron (MLP)~\cite{dehghani2008neural}, while Lee et al. used MLP as an estimator to capture the input-output relationship of the Stewart platform~\cite{sang1999estimation}. 
The Radial Basis Function (RBF) network was used in a hybrid robot for learning forward kinematics~\cite{kang2012learning}. 
Prado et al. applied baseline neural networks to solve the forward kinematics of a wearable parallel robot~\cite{prado2021artificial}. 

In this paper, we first design and fabricate a three-chain Revolute-Revolute-Universal-Revolute (3-RRUR) parallel robot to provide traction support for patients who suffer from cervical spondylosis. 
The use of the RRUR structure for each chain is motivated by the need to enlarge the workspace compared to an earlier related work that used a Revolute-Prismatic-Universal-Revolute (RPUR) structure~\cite{zhou2024design}.
The inverse kinematics is first solved analytically (Section~\ref{sec:analytical}) to generate data which are in turn used to train two distinct learning-based methods considered herein. 
Then, forward kinematics is estimated using the Koopman operator (Section~\ref{sec:koopman}) and a recurrent neural network (RNN; Section~\ref{sec:rnn}). 
Extensive testing in simulation and via physical experiments (Section~\ref{sec:results}) demonstrate the feasibility of both learning-based methods and offer insights as to when one method might be preferred over the other. 
This work provides a key step toward porting data-driven methods into robotic orthotics and establishes a basis for more Koopman operator-based methods to be investigated within this context. 
%
The contributions are summarized as follows:
\begin{itemize}
  \item[$\bullet$] A 3-RRUR parallel robot is designed to support patients who suffer from cervical spondylosis. The physical and kinematics models are described in detail.
  \item[$\bullet$] Two data-driven methods, the Koopman operator and an RNN, and an analytical method, are deployed to solve the forward kinematics in simulation. To the best of the authors' knowledge, this is the first attempt to utilize the Koopman operator on a 3-RRUR parallel robot.
  \item[$\bullet$] A physical prototype of the parallel robot is fabricated and tested in the real world with the analytical method.
\end{itemize}

\section{Analytical Kinematic Analysis}\label{sec:analytical}

\subsection{Physical Structure of the 3-RRUR Parallel Robot}

\begin{figure}[!t]
\vspace{6pt}
\centering
\includegraphics[width=0.8\linewidth]{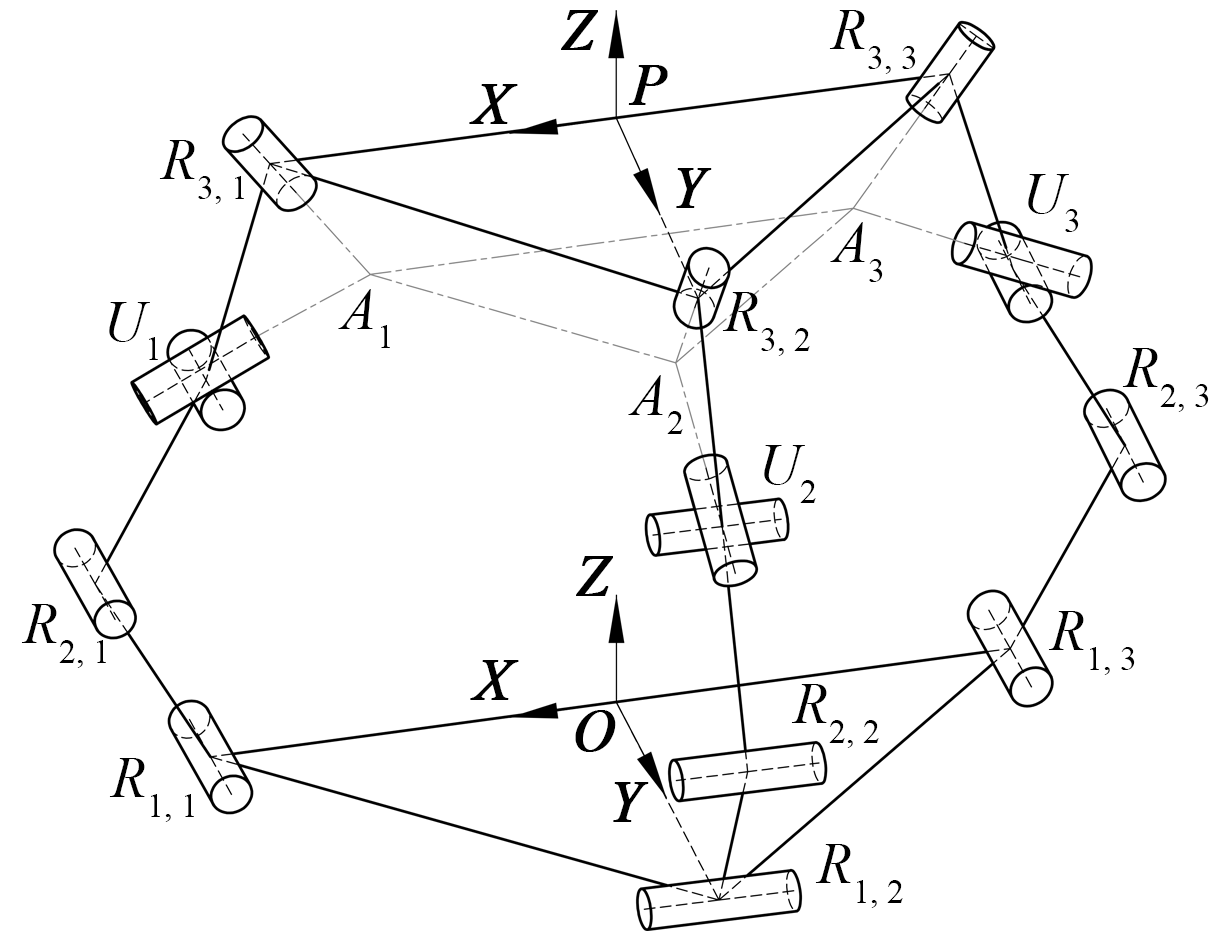}
\caption{Diagram of the developed 3-RRUR parallel robot.}
\label{fig:diagram_3RRUR}
\end{figure}

\begin{figure}[!t]
    \centering
    \includegraphics[width=0.6\linewidth]{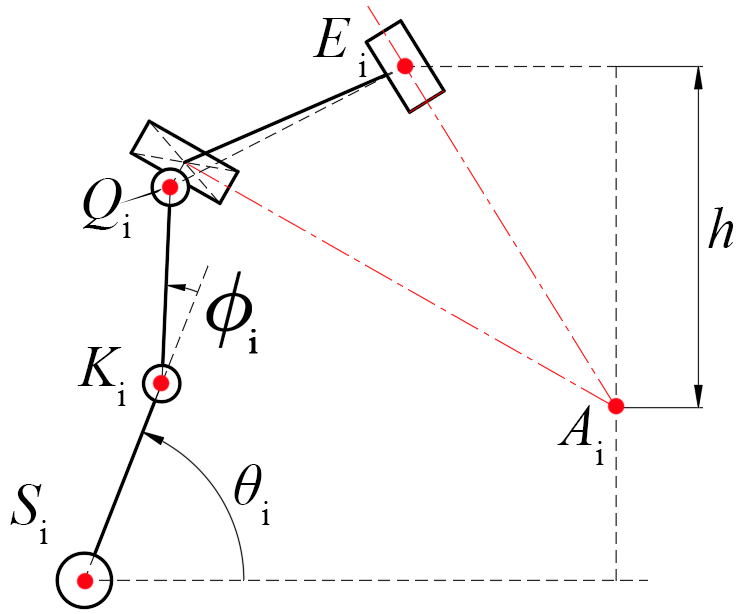}
    \caption{Chain structure diagram of 3-RRUR parallel robot}
    \label{fig:chain_structure}
\end{figure}

The physical structure of the developed parallel robot is shown in Figure~\ref{fig:diagram_3RRUR}. 
The robot is composed of three chains. 
A kinematic description of each chain is shown in Figure~\ref{fig:chain_structure}. 
The joint centers in each chain are represented as $S_{i}$, $K_{i}$, $Q_{i}$, and $E_{i}$, $(i=1,...,3)$ for joints $R_{1,i}$, $R_{2,i}$, $U_{i}$, and $R_{3,i}$, respectively. 
Each chain has five DoFs and provides one constraint to the end-effector frame, $P$. 
In total, the end-effector frame has three independent DoFs, which can be controlled by three rotary motors located on the first revolute joint of each chain, attaching them to the base plate characterized by frame $O$.
In each chain, the axes of the first two revolute joints and one axis of the universal joint are chosen to be perpendicular to the plane in which the chain operates. 
The intersection of the other axis of the $U_i$ joint and the axis of the third revolute joint $R_{3,i}$ is denoted by point $A_i$. 
Intersection points $A_{1}$, $A_{2}$ and $A_{3}$ form an intermediate plane. 
With the particular choice of geometric constraints, $A_1$ and $A_3$ stay in the same plane ${XZ}$ (${^{O}A_{1y}}=0$, $^{O}A_{3y}=0$), while $A_2$ stays in plane ${YZ}$ ($^{O}A_{2x}=0$).

\subsection{Inverse Kinematics}
\begin{figure}[!t]
\vspace{3pt}
    \centering
    \includegraphics[width=0.8\linewidth]{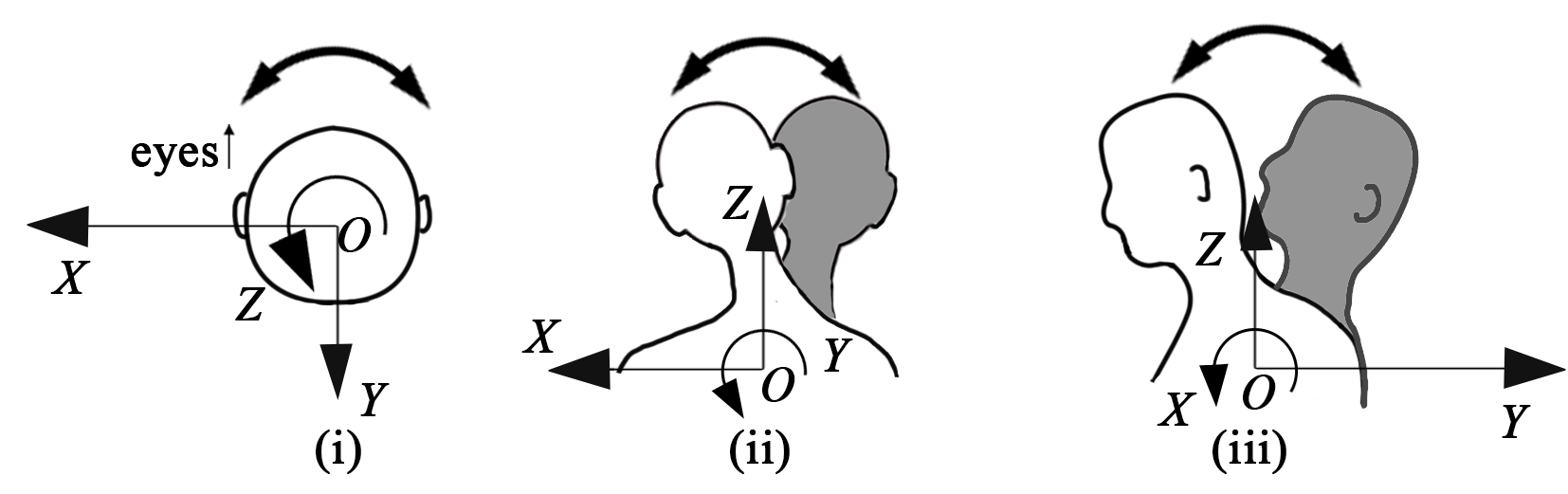}
    \vspace{-9pt}
    \caption{Description of three orientations of the head-neck aligned with the fixed base frame shown in Figure~\ref{fig:diagram_3RRUR}: (i) axial rotation, (ii) lateral bending, (iii) flexion and extension.}
    \label{fig:head_pose}
    \vspace{-12pt}
\end{figure}

Since the end-effector has three DoFs that can be independently controlled, we choose those DoFs that are directly related to head-neck motion as input to inverse kinematics. 
The three main head orientations to be supported by the mechanism are shown in Figure~\ref{fig:head_pose}. 
The chosen DoFs are the z-translation, $z_p$, flexion/extension, $\beta$, and lateral bending, $\gamma$. 
The rest of the DoFs of the end-effector are accommodated based on the constraints once the aforementioned actuated DoFs obtain valid values. 
Here we choose Space-three 3-1-2 rotation matrix~\eqref{eqn:rotation_sequence} to express the rotation between the base frame $O$ and end-effector frame $P$, that is
\begin{equation}
    \label{eqn:rotation_sequence}
    R_O^P =
    \begin{bmatrix}
        s_{\alpha} s_{\beta} s_{\gamma} + c_{\gamma} c_{\alpha} &
        c_{\alpha} s_{\beta} s_{\gamma} - c_{\gamma} s_{\alpha} &
        c_{\beta} s_{\gamma} \\
        s_{\alpha} c_{\beta} &
        c_{\alpha} c_{\beta} &
        -s_{\beta} \\
        s_{\alpha} s_{\beta} c_{\gamma} - s_{\gamma} c_{\alpha} &
        c_{\alpha} s_{\beta} c_{\gamma} + s_{\gamma} s_{\alpha} &
        c_{\beta} c_{\gamma} \\
        \end{bmatrix}\;.
\end{equation}

Each coordinate of $A_i$ in the base frame can be written as
\begin{equation}
    \label{eqn: Aio}
    ^{O}A_{i} = [x_p, y_p, z_p]^T + R_O^P (^{P}A_{i})\;,
\end{equation}
where $[x_p, y_p, z_p]^T$ represents the origin of the end-effector frame expressed in the base frame. 
Further, we have $^{P}A_{1}=[r_A, 0, -h]^T$, $^{P}A_{2}=[0, r_A, -h]^T$, and $^{P}A_{3}=[-r_A, 0, -h]^T$. 
The axial rotation of the end-effector is equal to zero. 
Then,
\begin{equation}
    \label{eqn:A1}
    ^{O}A_{1} = 
    \begin{bmatrix}
    x_p \\ y_p \\ z_p    
    \end{bmatrix} 
    +
    R_O^P
    (^{P}A_{1})
    =
    \begin{bmatrix}
    x_p + r_A c_{\gamma} - h c_{\beta} s_{\gamma} \\
    y_p + h s_{\beta}\\ 
    z_p - r_A s_{\gamma} - h c_{\beta} c_{\gamma}   
    \end{bmatrix}
\end{equation}
\begin{equation}
    \label{eqn:A2}
    ^{O}A_{2} = 
    \begin{bmatrix}
    x_p \\ y_p \\ z_p    
    \end{bmatrix} 
    +
    R_O^P
    (^{P}A_{2})
    =
    \begin{bmatrix}
    x_p + r_A s_{\beta} s_{\gamma}- h c_{\beta} s_{\gamma} \\
    y_p + r_A c_{\beta} + h s_{\beta}\\ 
    z_p + r_A s_{\beta} c_{\gamma} - h c_{\beta} c_{\gamma}
    \end{bmatrix}
\end{equation}
\begin{equation}
    \label{eqn:A3}
    ^{O}A_{3} = 
    \begin{bmatrix}
    x_p \\ y_p \\ z_p    
    \end{bmatrix} 
    +
    R_O^P
    (^{P}A_{3})
    =
    \begin{bmatrix}
    x_p - r_A c_{\gamma} - h c_{\beta} s_{\gamma} \\
    y_p + h s_{\beta}\\ 
    z_p + r_A s_{\gamma} - h c_{\beta} c_{\gamma}   
    \end{bmatrix}
\end{equation}

where $h$ and $r_A$ are design constants of the parallel robot; $h$ stands for the distance between plane $A_1A_2A_3$ and plane $E_1E_2E_3$, while $r_A$ stands for the radius formed by circle $A_1A_2A_3$. 
Considering the geometry constraints, the x-translation, $x_p$, and y-translation, $y_p$, are expressed as
\begin{equation}
\label{eqn:x_p}
    x_p = h c_{\beta} s_{\gamma} - r_A s_{\beta} s_{\gamma}\;,
\end{equation}
\begin{equation}
\label{eqn:y_p}
    y_p = -h s_{\beta}\;.
\end{equation}

To solve for the joint variables $\theta_i$ and $\phi_i$ for each chain, we use
\begin{equation}
    \label{eqn:QE}
    ||^{O}Q_{i}(\theta_i, \phi_i) - {^{O}E_{i}}||^2 = l_{Q_{i}E_{i}}^2\;,
\end{equation}
\begin{equation}
    \label{eqn:QA}
    ||^{O}Q_{i}(\theta_i, \phi_i) - {^{O}A_{i}}||^2 = l_{Q_{i}A_{i}}^2\;.
\end{equation}
where $l_{Q_iE_i}$ and $l_{Q_iA_i}$ are design constants of the parallel robot, and $^{O}Q_{i}(\theta_i, \phi_i)$, $i=1,..., 3$, represent the joint coordinates in the base frame.

\subsection{Forward Kinematics}
Three sets of equations (each containing two functions $f_1$ and $f_2$) can be found after resolving the inverse kinematics. 
We can write these as follows (detailed expressions of those equations can be found in the appendix).
\begin{equation}
    f_1(\theta_i, \phi_i, z_p, \beta, \gamma)=0\;\textrm{and}\;
    f_2(\theta_i, \phi_i, z_p, \beta, \gamma)=0\;.
\end{equation}
The unknown variables now are $\phi_1$,  $\phi_2$, $\phi_3$, $z_p$, $\beta$, and $\gamma$. Concerning the safety during the usage of the mechanism, we do not expect the mechanism to have ``inward" or ``elbow up" postures ($\theta_i<90^\circ$). 
Therefore, we are only interested in the solutions with $\theta_i>90^\circ$ from the inverse kinematics. 
Thus, those solutions from inverse kinematics should match the forward kinematics. 
Further, $\theta_i$ joint values obtained from the inverse kinematics, when sent as input to forward kinematics, should produce the same end-effector pose as in the inverse kinematics. 
To circumvent the complex procedure of solving the forward kinematics directly, data-driven approaches are introduced in the following sections.

\section{Koopman-based Kinematics Estimation}\label{sec:koopman}

\subsection{Koopman Operator Overview}
The Koopman operator efficiently learns the system's dynamics by lifting the original states to a latent space and capturing the flow of the lifted states (observables) with a finite number of linear functions. 
For an unforced system 
$x_{t+1}=f(x_t)$ 
with flow $\Phi(t,x)$ and complex-valued observables $\varphi\in\mathcal{F}$, the continuous time Koopman operator $\mathcal{K}:\mathcal{F}\to  \mathcal{F}$ is defined as 
$(\mathcal{K}\varphi)(\cdot)=\varphi \circ \Phi(t,\cdot)$. 
Being linear, the Koopman operator can be characterized by its eigenvalues and eigenfunctions. A function $\phi$ is an eigenfunction of $\mathcal{K}$ if  
$(\mathcal{K}\varphi)(\cdot)=e^{\lambda t}\phi(\cdot)$, 
with $\lambda$ being the corresponding eigenvalue~\cite{nathan2018applied}. 
The time-varying observable $ \psi(t, x) = \mathcal{K}\varphi(x)$ is the solution of the PDE~\cite{surana2016koopman}, 
\begin{equation}\label{e8}
\begin{array}{c}
\frac{\partial \psi}{\partial t}= \mathcal{L}_f \psi \\
\psi(0, x) = \varphi(x_0)
\end{array}\;,
\end{equation}
where $x_0$ is the initial condition for the unforced system and $\mathcal{L}_f$ is the Lie derivative with respect to $f$.

A vector-valued observable $\mathrm{g}(\cdot)$ may be expressed in terms of Koopman eigenfunctions $\phi_i$ as 
$\mathrm{g}(\cdot) = \sum_{i=1}^{\infty }\phi_i(\cdot)v_i$, 
where $v_i$ denote 
the Koopman modes. 
Koopman modes are obtained from the projection of the observable on the span of Koopman eigenfunctions. 
Then, 
\begin{equation}\label{eqn:k_formula}
    \mathcal{K}\mathrm{g}(\cdot) =\mathcal{K} \sum_{i=1}^{\infty }\phi_i(\cdot)v_i =  \sum_{i=1}^{\infty } \mathcal{K}\phi_i(\cdot)v_i = \sum_{i=1}^{\infty } \lambda_i\phi_i(\cdot)v_i\;.
\end{equation}

\subsection{Implementation}

Herein, the Koopman operator is utilized to learn the mapping from the motor angles, $\theta_1, \theta_2, \theta_3$, to end-effector angles $\beta$ and $\gamma$, and translation $z_p$. 
The x- and y-translations can be then computed by~\eqref{eqn:x_p} and~\eqref{eqn:y_p}. 
Suppose there are $M$ sets of motor angles and corresponding end-effector poses. 
Let $X=[x_1, x_2, ..., x_M]$ denote the $M$ motor angle states, where $x_j = [\theta_{j1} \quad \theta_{j2} \quad \theta_{j3}]^T, j=1,2,...,M$, and $Y=[y_1, y_2, ..., y_M]$ denote the corresponding $M$ end-effector poses, where $y_j = [\beta_{j} \quad \gamma_{j} \quad z_{pj}]^T, j=1,2,...,M$. 
For the lifting function design, which plays a crucial part in approximating the true system propagation model, we set
\begin{equation}
\begin{split}
    \Psi(x_j) = &  (1, \theta_{j1}, \theta_{j1}^2, \sin{\theta_{j1}}, \cos{\theta_{j1}}) ~\otimes \\ {} &  (1, \theta_{j2}, \theta_{j2}^2, \sin{\theta_{j2}}, \cos{\theta_{j2}}) ~\otimes \\ {} &  (1, \theta_{j3}, \theta_{j3}^2, \sin{\theta_{j3}}, \cos{\theta_{j3}})\;,
\end{split}
\end{equation}
where $\otimes$ stands for the Kronecker product. 
Note that this is a minimal dictionary. 
A larger dictionary consisting of more complex operations may increase prediction accuracy at the expense of increasing complexity. 
Construction of the dictionary is out of the scope of this paper; we refer the interested reader to~\cite{shi2021acd} for a detailed analysis of how to construct such a dictionary based on an underlying system's kinematic information. 
The Koopman operator can be numerically approximated from data using EDMDc~\cite{korda2018linear}. 
It entails the minimization of the total residual between the prediction and ground truth in the Koopman state, i.e.
\begin{equation}\label{eq:minim}
    J=\frac{1}{2}\sum_{j=1}^M(\Psi(y_j)-K\Psi(x_j))^2\;.
\end{equation}

This least-squares problem~\ref{eq:minim} can be solved by truncated singular value decomposition (SVD) as
\begin{equation}\label{eqn:K_formulation}
    K \triangleq G^\dag A\;,
\end{equation}
where 
\begin{gather}
    G=\frac{1}{M} \sum_{j=1}^M \Psi^*(x_j)\Psi(x_j) \\
    A=\frac{1}{M} \sum_{j=1}^M \Psi^*(x_j)\Psi(x_{j+1}),
\end{gather}
with $\dag$ denoting the pseudoinverse and $*$ denoting the conjugate transpose operations, respectively.

With $K$ in \eqref{eqn:K_formulation}, the following equations are obtained
\begin{equation} \label{eqn:k_decomposition}
\left\{  
        \begin{array}{l}
            v_n=(w_n^*B)^T\\  
            \lambda_n\xi_n = K\xi_n, \\  
            \varphi_n = \Psi \xi_n
        \end{array}
\right.      
\end{equation}
where $\xi_n$ is the $n$-th eigenvector, $w_n$ is the $n$-th left eigenvector of $K$, and $B$ is the matrix of appropriate weighting vectors so that $x = (\Psi B)^T$. By plugging expressions~\eqref{eqn:k_decomposition} back to~\eqref{eqn:k_formula} we can then describe the evolution of the system using the estimated Koopman operator.

\section{RNN-based Kinematics Estimation} \label{sec:rnn}
\subsection{RNN Overview}
RNNs are a class of deep neural networks designed for sequential data processing and temporal modeling. 
Unlike feedforward networks, RNNs have a recurrent structure that allows information from previous time steps to be recorded and used in the current computation. 
This is achieved through feedback connections in the hidden layers, enabling RNNs to capture dependencies over time. 
The hidden state $h_t$ at time step $t$ is updated using \(h_t=\mathit{f}(W_h h_{t-1}+W_x x_t+b)\) where $\mathit{f}$ is a nonlinear activation function, $W_h$ and $W_x$ are weight matrices, $x_t$ is the input, and $b$ is the bias. 
In the context of solving the forward kinematics of parallel robots, RNNs offer a distinct advantage due to their ability to model the inherent dependencies between joint variables and the resulting end-effector poses. 
\subsection{Implementation}
\begin{figure}[!t]
\vspace{2pt}
    \centering
    \includegraphics[width=0.99\linewidth]{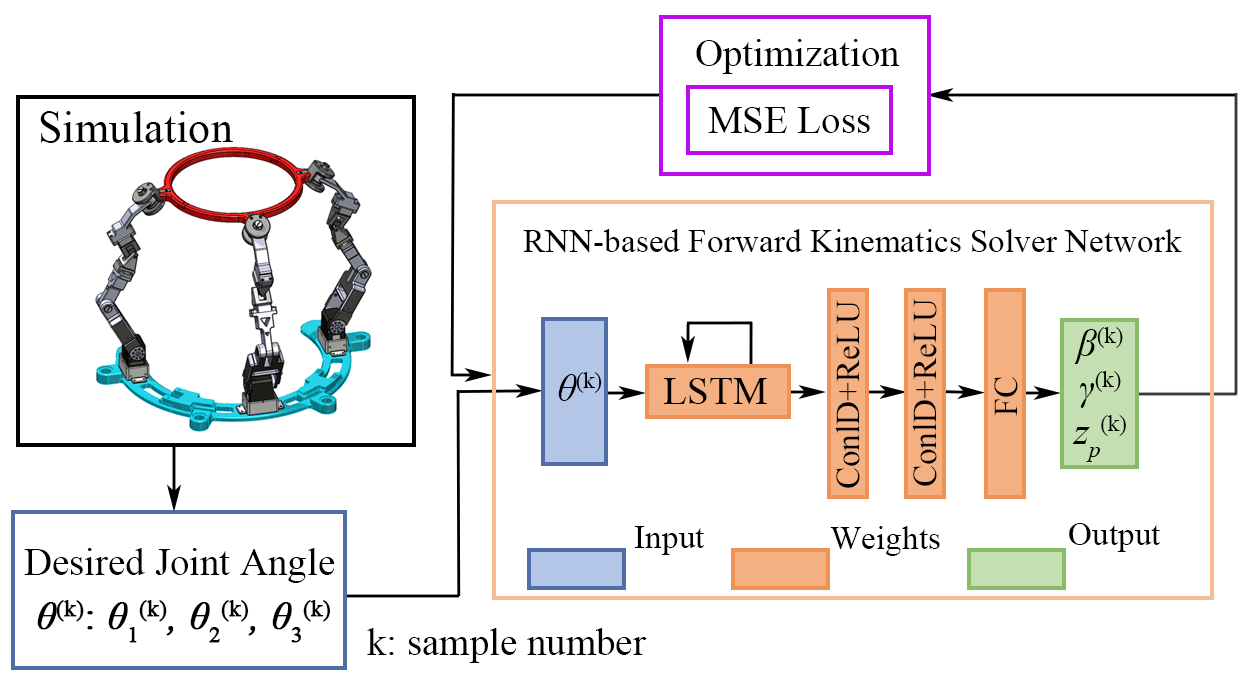}
    \vspace{-18pt}
    \caption{The RNN-based architecture considered herein.}
    \label{fig:rnn}
    \vspace{-12pt}
\end{figure}

The employed RNN architecture is shown in Figure~\ref{fig:rnn}. 
The {\it RNN-based Forward Kinematics Solver Network} module receives the three motor angles as inputs and outputs the flexion angle ($\beta$), lateral bending angle ($\gamma$), and the $z$-translation ($z_p$) of the end-effector. 
Training data were generated using the analytical inverse kinematic solver. 
The inputs and outputs from the analytical inverse kinematic solver were flipped and fed into the network. 


\section{Experimental Testing and Results}\label{sec:results}

Validation and testing were formulated into two parts; simulations and physical experiments. 
Simulations aim to assess the accuracy of predicting the end-effector pose from the active joint variables using the Koopman operator-based and RNN-based estimators. 
To establish a ground truth for validation, a trajectory of end-effector positions and orientations was generated from inverse kinematics. 
This trajectory serves as a benchmark to evaluate the performance of both methods. 
%
A dataset consisting of 248,790 samples was created for training. 
%
A distinct set of 802 samples was used for testing. 
Simulations with the Koopman Operator and the RNN-based estimator (including training and testing) were performed on an Intel Core i7-11800H processor @2.3GHz with RTX3060 Mobile GPU with 6GB of VRAM. 

\begin{figure}[!t]
\vspace{0pt}
    \centering
    \includegraphics[width=0.90\linewidth]{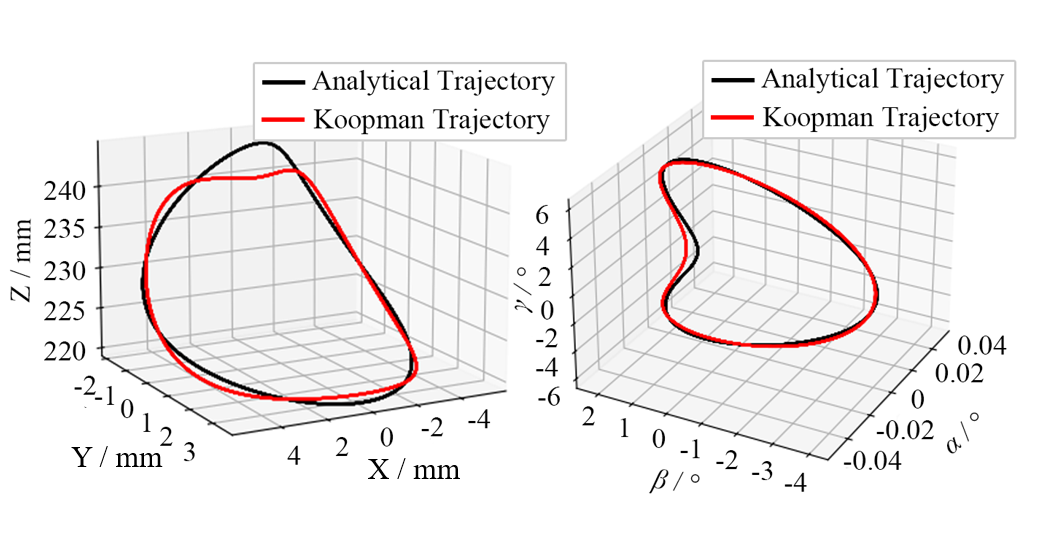} 
    \vspace{-18pt}
    \caption{Predicted end-effector position (left) and orientation (right) in simulation using the Koopman operator-based estimator.}
\label{fig:Koopman_prediction}
\vspace{-8pt}
\end{figure}

\begin{figure}
    \centering
\includegraphics[width=0.99\linewidth]{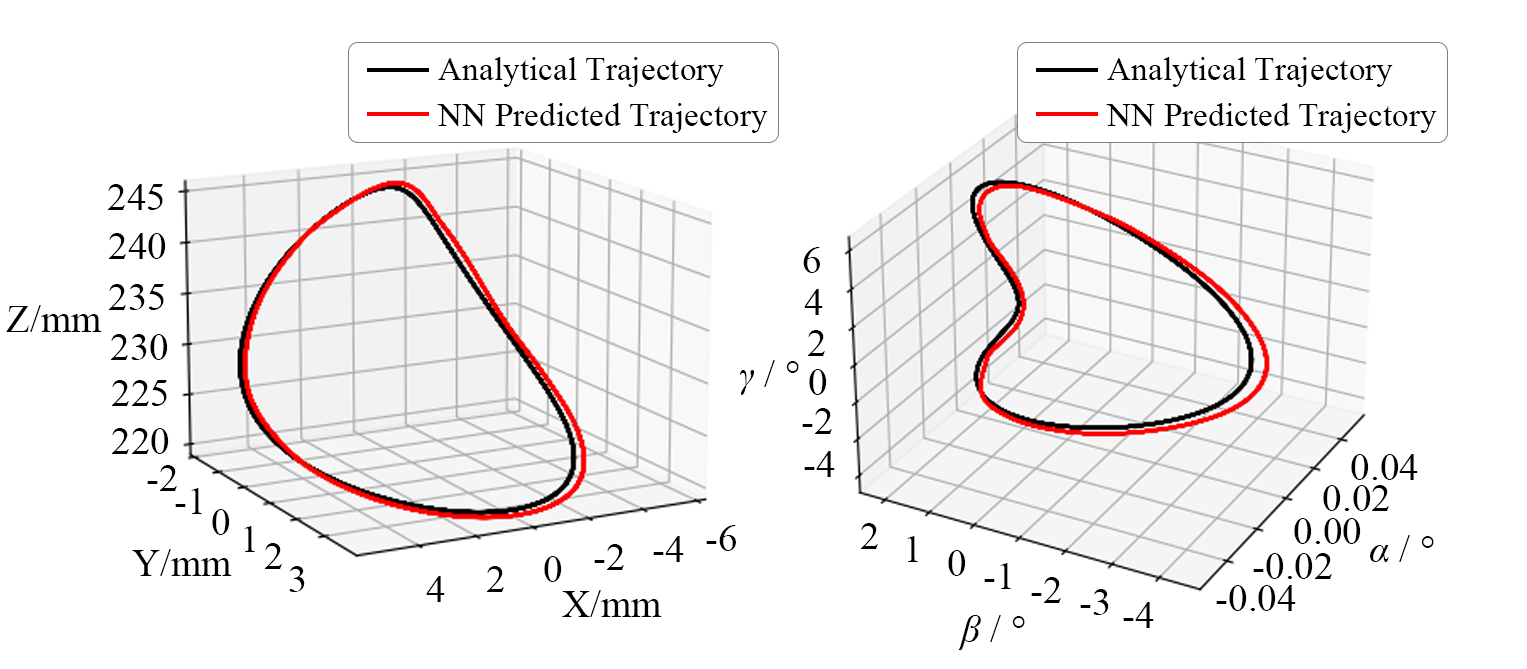}
\vspace{-22pt}
    \caption{Predicted end-effector position (left) and orientation (right) in simulation using the RNN-based estimator.}
    \label{fig:nn_prediction}
    \vspace{-12pt}
\end{figure}

Results for the Koopman operator-based and RNN-based estimator are shown in Figures~\ref{fig:Koopman_prediction} and~\ref{fig:nn_prediction}, respectively. 
%
%
%
It takes just 28.08 sec to train the Koopman operator. 
The average MSE in predicted trajectory positions is 3.24 (base units in mm), while the average MSE in predicted trajectory orientations is only $0.06$ (base units in deg). 
%
The RNN was trained for 50 epochs with a total training time of 455.55 sec. 
The model achieved a trajectory translation MSE of $0.0062$ (base units in mm) and orientation MSE of $0.0035$ (base units in deg). 
Additionally, the testing performance yielded an $\mathbf{R}^2$-score of 0.9974, indicating a high level of accuracy in the model's predictions. 
Overall, both methods' feasibility was shown in the simulated testing.

When comparing the two methods, we observe a notable trade-off between accuracy and computational efficiency. 
The RNN-based method achieves a significantly lower trajectory prediction error compared to that estimated via the Koopman operator.
However, this improved performance comes at the cost of computational efficiency since the RNN-based method requires approximately 10 times more training time to reach this level of accuracy.  
This highlights a critical consideration in selecting the appropriate method based on the specific requirements of accuracy and computational efficiency in practical applications. 
Deviations from the desired trajectory were observed in the predicted position (Figure~\ref{fig:Koopman_prediction}). 
However, the overall trajectory form is maintained. 
In addition, the prediction accuracy depends to a certain extent on the selection of the Koopman dictionary; a more extended dictionary could help rectify this. 

\begin{figure}[!t]
\vspace{6pt}
\centering
\includegraphics[width=0.6\linewidth]{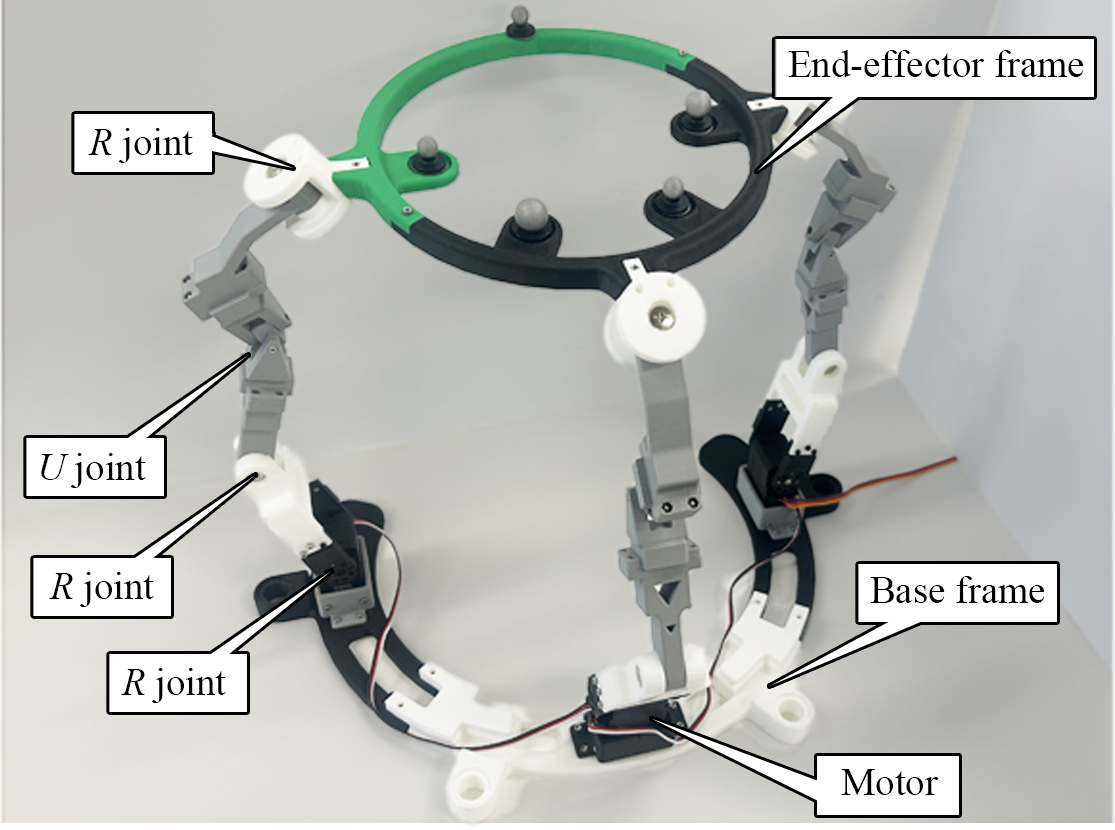}
\vspace{-8pt}
\caption{Fabricated prototype of the proposed 3-RRUR parallel robot. In each chain, the first $R$ joint that is connected to the base frame is actuated via a servomotor, whereas all other joints are passive and forced to move as determined by the geometry constraints.}
\label{fig:prototype}
\end{figure}

Physical experiments considered the same ground truth trajectory as in the simulation testing. 
The latter was applied to the fabricated 3-RRUR parallel robot prototype (Figure~\ref{fig:prototype}). End-effector frame $E_1E_2E_3$, base frame $S_1S_2S_3$, and link $S_iK_iQ_iE_i$ $(i=1,...,3)$ are 3D-printed with PLA basic material (Bambu Labs). 
Joints $S_i$ are formed by LX16A bus servomotors. 
Joints $K_i$, $Q_i$, $E_i$ are revolute joints formed by binding barrels with 304 stainless steel phillips head and deep groove ball bearings. 
The physical trajectory data were captured by an 8-camera Optitrack motion capture system at 100Hz. 
Key dimensions of the prototype include: $r_A = 85.04$\,mm, $r_S = 141$\,mm, $r_E = 120$\,mm, $h = 48.51$\,mm, $l_{Q_iA_i} = 66.56$\,mm, $l_{Q_iE_i} = 85.06$\,mm, $l_{K_iS_i} = 83.925$\,mm, $l_{K_iQ_i} = 83.925$\,mm, $l_{E_iA_i} = 59.76$\,mm.

\begin{figure}[!t]
\centering
\includegraphics[width=0.99\linewidth]{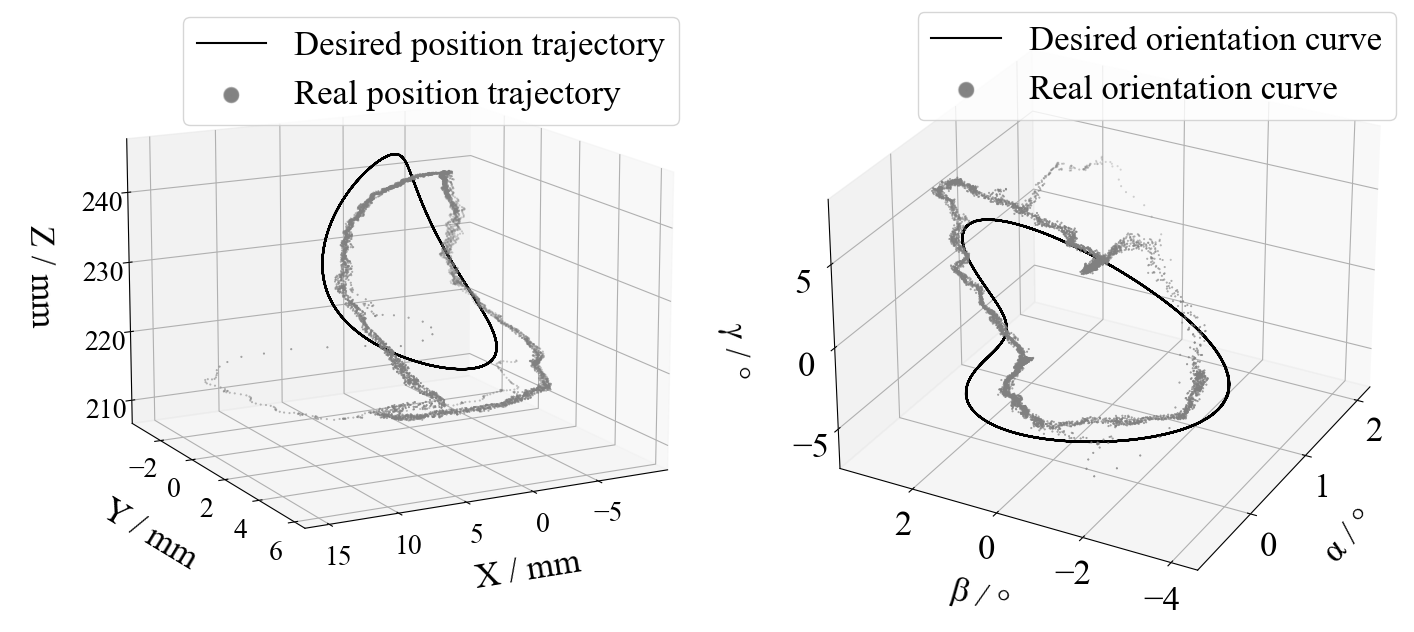}
\vspace{-22pt}
    \caption{Comparison between desired trajectory and the experimentally-obtained one collected via motion capture.}
    \label{fig:real_exp}
    \vspace{-12pt}
\end{figure}

Results of the captured trajectory and desired trajectory are shown in Figure~\ref{fig:real_exp}. 
The overall trajectory trend was followed by the feedforward trajectory computed analytically. 
However, deviations and, at times, large offsets were observed as well. 
Two likely sources caused these errors. 
First, differences in assumptions used in the mathematical model and the real physical prototype (for instance the geometry constraints of $A_i$) cannot be fully achieved because of the deformation of each link. 
Second, fabrication errors, especially those causing friction in the revolute joints and collision of links on each chain $S_iK_iQ_iE_i$, can all contribute to cumulative errors in the end-effector motion.

\section{Conclusion}
A 3-RRUR parallel robot was designed as a means to provide traction support for patients who suffer from cervical spondylosis. 
To address the challenge in forward kinematics analysis associated with the parallel mechanism structure, an analytical method for solving inverse kinematics, which can often be solved directly in contrast to forward kinematics, was studied to generate a dataset for data-driven model training. 
Then, a Koopman operator-based method and an RNN-based method were introduced to resolve the forward kinematics of the robot. 
The prediction accuracy from the RNN-based estimator is noticeably higher than that of the Koopman operator, while the Koopman operator can be much more quickly and easily deployed and updated online, without the need for large computational resources. 
Lastly, a fabricated prototype of the proposed 3-RRUR parallel robot was designed to validate the results of the same trajectory presented in the two data-driven methods. 
This work offers a basis to support porting data-driven methods into robotic orthotics. 
Future directions of research include
online model learning via the Koopman operator to control the mechanism, as well as design and fabrication improvements to make the motion of the physical prototype smoother. 

\section*{Acknowledgment}
The authors would like to thank Dr. Zhichao Liu for offering his assistance with the experiments.

\section*{Appendix}
We list below the closed-form expressions for the forward kinematics of the developed mechanism. 
Let $r_E$ be the diameter of the circle spanned by points $E_1$, $E_2$, $E_3$ and $r_S$ the diameter of the circle spanned by points $S_1$, $S_2$, $S_3$. 
The two kinematic constraints in chain $S_1K_1Q_1E_1$ are 

\begin{equation}
    \label{eqn:Q1E1}
    f_1: (a_{11} + z_{11}) ^ 2 + (a_{21} - z_{21})^2 = a_{31}\;,
\end{equation}
\begin{equation}
    \label{eqn:Q1A1}
    f_2: (a_{41} + z_{11}) ^ 2 + (a_{51} - z_{21})^2 = a_{61}\;,
\end{equation}
%
where $a_{11} = x_p + r_E c_{\gamma} - r_S$, 
$a_{21} = z_p - r_E s_{\gamma}$
$a_{31} = l_{QE}^2 - y_{E}^2$, 
$a_{41} = x_p + r_A c_{\gamma} - h c_{\beta} s_{\gamma} - r_S$, 
$a_{51} = z_p - r_A s_{\gamma} - h c_{\beta} c_{\gamma}$, 
$a_{61} = l_{Q_1A_1}^2$, 
$z_{11} = l_{K_1S_1} c_{\theta_1} + l_{K_1Q_1} c_{\theta_1 + \phi_1}$, and
$z_{21} = l_{K_1S_1} s_{\theta_1} + l_{K_1Q_1} s_{\theta_1 + \phi_1}$. 

The two kinematic constraints in chain $S_2K_2Q_2E_2$ are
\begin{equation}
    \label{eqn:Q2E2}
    f_1:(a_{12} + z_{12})^2 + (a_{22} - z_{22})^2 = a_{32}\;,
\end{equation}
\begin{equation}
    \label{eqn:Q2A2}
    f_2:(a_{42} + z_{12})^2 + (a_{52} - z_{22})^2 = a_{62}\,,
\end{equation}
where
$a_{12} = y_p + r_E c_{\beta} - r_S$,
$a_{22} = z_p + r_E s_{\beta} c_{\gamma}$,
$a_{32} = l_{Q_2E_2}^2 - (x_p + r_E s_{\beta} s_{\gamma})^2$,
$a_{42} = r_A c_{\beta} - r_S$,
$a_{52} = z_p + r_A s_{\beta} c_{\gamma} - h c_{\beta} c_{\gamma}$,
$a_{62} = l_{Q_2A_2}^2$,
$z_{12} = l_{K_2S_2} c_{\theta_2} + l_{K_2Q_2} c_{\theta_2 + \phi_2}$, and
$z_{22} = l_{K_2S_2} s_{\theta_2} + l_{K_2Q_2} s_{\theta_2 + \phi_2}$.

The two kinematic constraints in chain $S_3K_3Q_3E_3$ are
\begin{equation}
    \label{eqn:Q3E3}
    f_1:(a_{13} - z_{13})^2 + (a_{23} - z_{23})^2 = a_{33}\;,
\end{equation}
\begin{equation}
    \label{eqn:Q3A3}
    f_2:(a_{43} - z_{13})^2 + (a_{53} - z_{23})^2 = a_{63}\;,
\end{equation}
where
$a_{13} = x_p - r_E c_{\gamma} + r_S$,
$a_{23} = z_p + r_E s_{\gamma}$,
$a_{33} = l_{Q_3E_3}^2 - y_{E}^2$, 
$a_{43} = x_p - r_A c_{\gamma} - h c_{\beta} s_{\gamma} + r_S$,
$a_{53} = z_p + r_A s_{\gamma} - h c_{\beta} c_{\gamma}$,
$a_{63} = l_{Q_3A_3}^2$,
$z_{13} = l_{K_3S_3} c_{\theta_3} + l_{K_3Q_3} c_{\theta_3 + \phi_3}$, and
$z_{23} = l_{K_3S_3} s_{\theta_3} + l_{K_3Q_3} s_{\theta_3 + \phi_3}$.

\bibliographystyle{IEEEtran}
\bibliography{ref}

\end{document}